\documentclass[10pt,letterpaper]{article}
\usepackage[utf8]{inputenc} 
\usepackage[T1]{fontenc}    
\usepackage{hyperref}       
\usepackage{url}            
\usepackage{booktabs}       
\usepackage{amsmath}
\usepackage{amsfonts}       
\usepackage{nicefrac}       
\usepackage{microtype}      
\usepackage[normalem]{ulem}

\usepackage{graphicx,graphics,subcaption}
\usepackage{tikz,pgf,pgfplots}
\usepackage{authblk}

\usepackage{todonotes}
\presetkeys{todonotes}{color=blue!20}{}

\title{Understanding Learning Dynamics of Binary Neural Networks via Information Bottleneck}

\author{
Vishnu Raj\footnote{ee14d213@ee.iitm.ac.in}, 
Nancy Nayak\footnote{ee17d408@smail.iitm.ac.in} and 
Sheetal Kalyani\footnote{skalyani@ee.iitm.ac.in} \\
Department of Electrical Engineering, IIT Madras. \\
}

\begin{document}
    \maketitle
    
    \begin{abstract}
    Compact neural networks are essential for affordable and power efficient deep learning solutions. Binary Neural Networks (BNNs) take compactification to the extreme by constraining both weights and activations to two levels, $\{+1, -1\}$. However, training BNNs are not easy due to the discontinuity in activation functions, and the training dynamics of BNNs is not well understood. In this paper, we present an information-theoretic perspective of BNN training. We analyze BNNs through the Information Bottleneck principle and observe that the training dynamics of BNNs is considerably different from that of Deep Neural Networks (DNNs). While DNNs have a separate empirical risk minimization and representation compression phases, our numerical experiments show that in BNNs, both these phases are simultaneous. Since BNNs have a less expressive capacity, they tend to find efficient hidden representations concurrently with label fitting. Experiments
    in multiple datasets support these observations, and we see a consistent behavior across different activation functions in BNNs.
\end{abstract}
    
    \section{Introduction}\label{intro}
The recent success of machine learning and artificial intelligence, to a major share, can be attributed to the developments in Deep Neural Networks (DNNs). Powered by superior function approximation capabilities \cite{cybenko1989approximation} and the availability of increased computational power, DNNs often surpass human performance \cite{he2015delving,nassif2019speech,mnih2015human} in complex real-world problems. Computational requirements of DNNs are often criticized for being high making it one of the major hindrances in the widespread adoption of deep learning solutions in daily life. Extending the advancements in DNN to cater to the requirements of general public warrants the need for methods that can efficiently run DNNs in devices with low computational as well as power requirements. This leads to the development of strategies for the compactification of neural networks and enables them to be used under small compute infrastructure with low power consumption. Major steps towards this include pruning of trained neural networks \cite{blalock2020state}, arbitrary quantization of model parameters \cite{nogami2019optimizing}, knowledge distillation \cite{hinton2015distilling}, etc. To push quantization to its extreme, work in \cite{courbariaux2016binarized} introduced neural networks that use only two values, $\{-1, +1\}$, for both weights and activations.

Even though DNNs are widely successful in multiple scenarios, our understanding of why deep learning works is fairly limited. Recent works that tried to develop theories explaining why deep learning works approached the topic from multiple perspectives, including optimization loss landscape \cite{nguyen2018optimization}, random matrix theory \cite{pennington2017nonlinear}, thermodynamics \cite{alemi2018therml}, statistical mechanics, \cite{bahri2020statistical} and information theory \cite{tishby2015deep}. In the absence of a unified theory for explaining the success of deep learning, each of these perspectives merits further investigation and insights derived following each approach can broaden our understanding of deep learning. In this work, we study the learning dynamics of Binary Neural Networks (BNNs) from the perspective of the Information Bottleneck (IB) principle \cite{tishby2000information, tishby2015deep}.

The Information Bottleneck (IB) principle was introduced in \cite{tishby2000information} as a framework to extract relevant information from random variable $X$ (features) about another random variable $Y$ (labels). By viewing each step of processing the input feature as a trade-off between compression of information and keeping relevant statistics helpful in prediction, IB principle presents an information-theoretic interpretation for explaining the learning dynamics in a supervised learning system. \cite{tishby2015deep} laid the groundwork for applying IB principle to analyze deep learning by formulating the information flow from the input layer to the output layer as successive Markov chains of intermediate representations. In a follow-up work, \cite{shwartz2017opening} analyzed the information plane dynamics of each layer during training to obtain insights into the DNN training process. By using a neural network with tanh activation, it was shown that neural networks get trained in two phases: i) \emph{empirical risk minimization} (ERM) phase where the stochastic gradient descent (SGD) algorithm generates high SNR gradients, the loss rapidly decreases and ii) a much slower \emph{representation compression} (RC) phase where efficient representations for intermediate activations are learned. In the information plane, this can be easily visualized as a rapid increase in the the Mutual Information (MI) between input and intermediate representation in the first phase followed by a slow decrease of the same during the second phase. It is also claimed that DNNs learn to generalize over the input domain in the second phase of compression. However, \cite{saxe2019information} argued that these observations are artifacts of using a double-sided saturating non-linearity like tanh and does not generally hold. By using the ReLU activation function, which is a single-sided saturating non-linearity, \cite{saxe2019information} empirically showed that the different phases during training generally does not exist and generalization in DNNs takes place concurrently during training rather than in a separate compression phase.

BNNs are now widely seen as a solution for the cost-effective adoption of deep learning advancements.  Hence, an insight into the learning dynamics can help in the development of efficient optimizers targeted towards training BNNs. In this paper, we present the learning dynamics of BNNs from an IB perspective and discuss how the optimization process of BNNs is different from traditional DNNs. Our major contributions are listed below.
\begin{enumerate}
    \item For the first time, the learning dynamics of Binary Neural Network is studied from an information theory perspective.
    \item We observe that the learning dynamics of BNNs are different from that of DNNs in information plane. While DNNs follow a two-phase approach for learning and generalizing, our results show that BNN training proceeds with simultaneous label fitting and generalization.
\end{enumerate}

    \section{Training Binary Neural Networks}\label{training_bnn}
Binary Neural Networks are characterized by both weights and activation taking only values $\{-1, +1\}$. This reduces the hardware complexity at the inference phase, as all operations can be carried out using binary primitives. During the training phase, however, high precision hardware is used to store the real weight counterpart values to enable backpropagation. These real-valued weights are binarized during the forward pass. As the SGD optimization procedure progresses, these full precision values are used for computing gradient updates for the parameters. 

\begin{figure}[!t]
    \centering
    \begin{subfigure}{0.32\linewidth}
        \resizebox{\linewidth}{!}{
        \begin{tikzpicture}
            \begin{axis}[
                width=8cm,height=6cm,
                ylabel=$g(x)$,xlabel=$x$,
                ymin=-1.25,ymax=1.25,xmin=-2,xmax=2,
                grid=major,
                legend pos=south east,
                legend cell align={left},
                legend style={fill opacity=0.6, draw opacity=1.0, text opacity=1.0, font=\small}
            ]
                \addplot[blue,smooth,domain=-2:0] {-1};
                \addlegendentry{Forward pass}
                
                \addplot[red,smooth,domain=-2:-1] {0};
                \addplot[red,smooth,domain=-1:+1] {1};
                \addplot[red,smooth,domain=+1:+2] {0};
                \addlegendentry{Backward pass}
                
                \addplot[blue,smooth,domain=0:+2] {+1};
            \end{axis}
        \end{tikzpicture}
        }
        \caption{STE \cite{courbariaux2016binarized}}
        \label{fig:stesign}
    \end{subfigure}
    \begin{subfigure}{0.32\linewidth}
        \resizebox{\linewidth}{!}{
        \begin{tikzpicture}
            \begin{axis}[
                width=8cm,height=6cm,
                ylabel=$g(x)$,xlabel=$x$,
                ymin=-2.25,ymax=2.25,xmin=-2,xmax=2,
                grid=major,
                legend pos=south east,
                legend cell align={left},
                legend style={fill opacity=0.6, draw opacity=1.0, text opacity=1.0, font=\small}
            ]
                \addplot[blue,smooth,domain=-2:0] {-1};
                \addlegendentry{Forward pass}
                
                \addplot[red,smooth,domain=-2:-1] {0};
                \addplot[red,smooth,domain=-1:+1] {2-2*abs(x)};
                \addplot[red,smooth,domain=+1:+2] {0};
                \addlegendentry{Backward pass}
                
                \addplot[blue,smooth,domain=0:+2] {+1};
            \end{axis}
        \end{tikzpicture}
        }
        \caption{Approximate sign \cite{liu2018bi}}
        \label{fig:approxsign}
    \end{subfigure}
    \begin{subfigure}{0.32\linewidth}
        \resizebox{\linewidth}{!}{
        \begin{tikzpicture}
            \begin{axis}[
                width=8cm,height=6cm,
                ylabel=$g(x)$,xlabel=$x$,
                ymin=-5.25,ymax=5.25,xmin=-2,xmax=2,
                grid=major,
                legend pos=south east,
                legend cell align={left},
                legend style={fill opacity=0.6, draw opacity=1.0, text opacity=1.0, font=\small}
            ]
                \addplot[blue,smooth,domain=-2:0] {-1};
                \addlegendentry{Forward pass}
                
                \addplot[red,smooth,domain=-2:+2] {(5 * (2 - 5*x*tanh(5*x/2)))/(1 + cosh(5*x))};
                \addlegendentry{Backward pass}
                
                \addplot[blue,smooth,domain=0:+2] {+1};
                
            \end{axis}
        \end{tikzpicture}
        }
        \caption{Swish sign  \cite{darabiregularized} $(\beta = 5.0)$}
        \label{fig:swishsign}
    \end{subfigure}
    \caption{Different activation functions for BNN models. Blue lines indicate the activation function $g(\cdot)$ in forward pass during training and inference. This is sign function. Red lines show the  functions used as differential $g'(\cdot)$ used during backpropagation.}
    \label{fig:bnn_activations}
\end{figure}
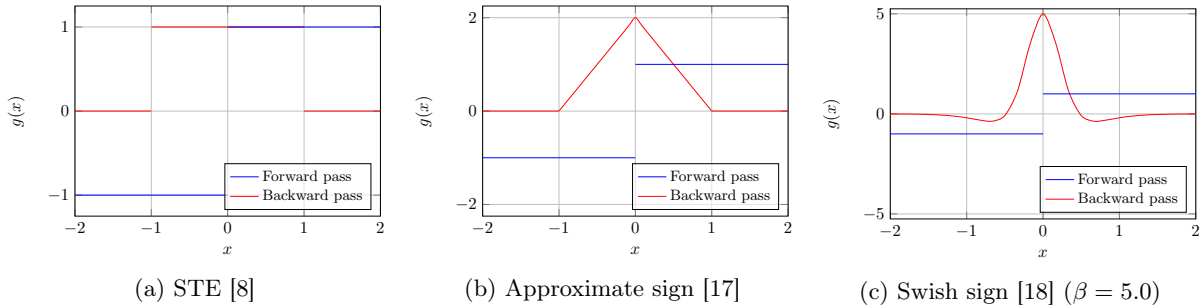

Training a BNN presents difficulty while computing the gradient updates for the parameters. The binary activation function $g(\cdot)$ in a BNN can be represented as,
\begin{align}
    g(x) = \begin{cases}
        -1 \qquad ; x \leq 0, \\
        +1 \qquad ; x > 0,
    \end{cases}
\end{align}
where $x$ is the result of the linear operation. The activation function in BNNs, the sign function, is neither differentiable nor continuous. However, backpropagation requires differentiable activation functions. In order to circumvent this problem, a widely accepted practice is to use a different function during backpropagation that is differentiable yet closely resembles the binary activation. In this work, we study the learning dynamics of BNNs with three popular binary activation functions.
\begin{enumerate}
    \item Straight-Through-Estimator (STE): STE-sign is used in \cite{courbariaux2016binarized} to train BNNs using backpropagation. The backward pass for STE is defined as,
    \begin{align}
        \frac{d}{dx}g(x) = \begin{cases}
            1 \qquad ; -1 \leq x \leq +1, \\
            0 \qquad ; \text{otherwise}.
        \end{cases}
    \end{align}
        
    \item Approximate sign: \cite{liu2018bi} introduced Approximate sign (approx-sign) function as  a tight approximation to the derivative of the non-differentiable sign function with respect to activation. The backward pass for approximate sign activation function is defined as,
    \begin{align}
        \frac{d}{dx}g(x) = \begin{cases}
            2 - 2|x| \qquad ; -1 \leq x \leq +1, \\
            0 \qquad \qquad ; \text{otherwise}.
        \end{cases}
    \end{align}
    
    \item Swish sign: \cite{darabiregularized} proposed swish sign activation as another close approximation for the sign function. The backward pass for swish sign activation function is defined as,
    \begin{align}
        \frac{d}{dx}g(x) = \frac{\beta\left( 2 - \beta \tanh\left( \frac{\beta x}{2}\right)\right)}{1 + \cosh{(\beta x)}}.
    \end{align}
\end{enumerate}
The backward pass of these activation functions are provided in Fig. \ref{fig:bnn_activations}. With these activations, the model weights for BNNs can be updated using SGD or a variant of it. A prevalent view that justifies the quantizing of real weights to obtain the binary model parameters is to consider the real weights as \emph{latent weights} \cite{courbariaux2016binarized}. In \cite{helwegen2019latent}, it is shown that this is not necessarily true and instead the real values parameters should be viewed as inertia parameters for each binary weights. Following this insight, the authors of \cite{helwegen2019latent} proposed the first optimizer that is specifically designed for BNN known as Binary Optimizer (Bop).

    \section{Information Bottleneck principle}
The aim of a deep neural network is to learn the optimal/coarsest representation of input features $x \in X$ with respect to the output labels $y \in Y$, where $X$ and $Y$ are dependent, and therefore have non-zero mutual information i.e. $I(X;Y)>0$. This optimal representation can be best characterized by minimal sufficient statistics, the coarsest partition of input space $X$. In the DNN setting, a sufficient statistics of $X$, denoted by $S(X)$, is the partition of $X$ that has all the information $X$ has on $Y$, i.e., $I(S(X);Y)=I(X;Y)$ \cite{cover2012elements}. The problem of finding the minimal sufficient statistics $T(X)$ becomes solving the constrained optimization problem:
\begin{equation}
    T(X) = \underset{S(X):I(S(X);Y)=I(X;Y)}{\arg \min} I(S(X);X).
\end{equation}
Since finding a minimal sufficient statistics is difficult in general distributions (except exponential families), \cite{tishby2000information} relaxed the optimization problem to find the approximate minimal sufficient statistics that captures as much $I(X; Y)$ as possible. This optimal trade-off between the compression of $X$ and the prediction of $Y$ was first coined by \cite{tishby2000information} as Information Bottleneck (IB) trade-off. Finding the compressed representation $T$ of $X$ becomes the minimization of the Lagrangian objective:
\begin{equation}
    \mathcal{L} = I(T;X) - \beta I(T;Y),
    \label{eqn:ib_obj}
\end{equation}
which implies that in a DNN we want to pass all of the information that $X$ has on $Y$ through a bottleneck formed by the compressed representation $T$. Here, $\beta$ is the Lagrange multiplier where $\beta = \infty$ implies no compression and vice versa. 

In \cite{tishby2015deep}, authors extended the notion of IB principle to deep neural networks. The structure of the DNN is reviewed as a Markov cascade of intermediate representations between input and output layers where each layer processes the output from the previous layer. Therefore, the information that is lost in one layer cannot be recovered in later layers. For example, let the set of hidden layers in a DNN is defined by $\mathcal{T}$ and $T_i$ denotes $i^{th}$ hidden layer and $i>j$, then according to data processing inequality (DPI) $I(Y;X)\geq I(Y;T_j) \geq I(Y;T_i)$. As discussed in Sec. \ref{intro}, authors in \cite{shwartz2017opening,saxe2019information} studied learning dynamics and characteristics of DNN in the information plane. In this work, we study the special characteristics of a BNN in the light of IB principle.

As BNNs have activations limited to only two values, this case is close to double saturating activation functions like tanh and hard-tanh (Refer Sec. \ref{App:act} in the appendix). We provide the information plane behavior of a deep feed-forward neural network with the above non-linear activations as the non-linearity after the dense layers and the results are given in Fig. \ref{fig:ibdnn_spherical_tanh} and Fig. \ref{fig:ibdnn_spherical_htanh} respectively.  \cite{darabiregularized} replaces traditional STE-sign activation with a sign-swish activation for better learning of the weights near $\{+1,-1\}$. The results with full precision sign-swish activation are provided in Fig. \ref{fig:ibdnn_spherical_signswish}. For training the real networks with tanh and hard-tanh activations, we use Adam optimizer with a learning rate of $0.0004$, and for DNN with sign-swish activation, we use Adam with learning rate $0.001$. To calculate the mutual information between input $X$ and output $Y$ with intermediate layers $T$, we choose binning (with $30$ bins) among all other mutual information estimators \cite{kraskov2004estimating,kolchinsky2017estimating, noshad2019scalable}.

We plot this in Fig. \ref{fig:ibdnn_spherical} for all the three activations in DNN and notice that the decrease in $I(T; Y)$ is prominent after $1000$ epochs for both the activations tanh and hard-tanh. We observe that the behavior is due to over-fitting. The same can be verified by checking the validation loss that jumps up around $1000$ epochs indicating a clear evidence of over-fitting. In the case of sign-swish, over-fitting is less and can be verified both in the information plane and the loss trend. 

\subsection{IB in BNNs}
In the information plane, the intermediate representation $T$ of a real-valued neural network can have high precision and hence can accommodate any shorter representation of the input. Therefore the information can flow from the input to the output without any hindrance. In contrast to the rate-distortion problem, where a coarser representation is required while keeping the distortion low, $I(T; Y)$ is used as a substitute for the distortion function that needs to be kept at a certain level for the correct prediction of $Y$ \cite{tishby2000information}. But as discussed in Sec. \ref{training_bnn}, different binary activations used in BNN constrain the activations to only two levels, and hence, the representation capability of $T$ is limited. Therefore, the free flow of complete information from the input layer to the output layer is suppressed. It is of immense interest to see the learning dynamics in case of BNN and analyze their behavior. To the best of our knowledge, analyzing the training dynamics of BNN with the help of the IB principle is not available in literature and is of immense interest.

    \section{Empirical Results}
In this section, we present numerical results exploring the applicability of the IB principle to BNNs. We study the learning dynamics of BNNs for the task of classification on a synthetic dataset and a real dataset, as described below.
\begin{enumerate}
    \item A synthetic dataset as studied in \cite{shwartz2017opening} is used. The input, $X$, is a $12$-dimensional vector of either $0$ or $1$ and is classified into two labels. There are $4096$ unique data points, out of $80\%$ is used as a train set and the rest as the validation set. The maximum $I(T;Y)$ is $1$ bit and maximum value for $I(T;X)$ is $12$ bits.
    \item Traditional MNIST dataset with $60000$ samples for training and $10000$ samples for validation. All pixel values are normalized to be in the range $[-1, +1]$. The maximum value for $I(T;Y)$ is $\log_2(10) \approx 3.22$ bits.
\end{enumerate}

For each of these datasets, fully connected BNN is trained for classification tasks with loss function as cross-entropy. BatchNorm layer is used between the linear operation and binary activation in each layer as done in \cite{courbariaux2016binarized}. The operations performed by a hidden layer in BNNs are i) linear operation using quantized weights (dense layer), ii) batch normalization without scaling, and iii) binary activation function. All results are averaged over $5$ independent runs.

Estimating MI between $X$, $T$, and $Y$ through empirical observations is challenging in DNNs as the activations can take real values. However, in the case of BNNs, the activations are limited to only two values. The only real values that get involved in the analysis of information plane dynamics are the values of input $X$ and output softmax layer $\hat{Y}$. Since these can be easily handled by binning, as shown in \cite{shwartz2017opening, saxe2019information}, we use a binning-based estimator for computing MI.

\subsection{Results on synthetic dataset}
\begin{figure}[!t]
    \centering
    \begin{minipage}{0.94\linewidth}
        \begin{subfigure}{0.32\linewidth}
            \includegraphics[width=\linewidth]{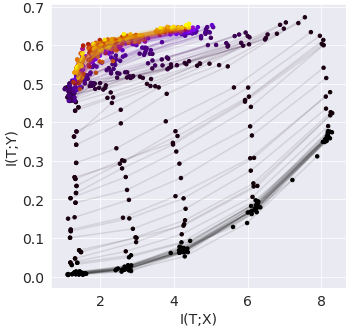}
        \end{subfigure}%
        \begin{subfigure}{0.32\linewidth}
            \includegraphics[width=\linewidth]{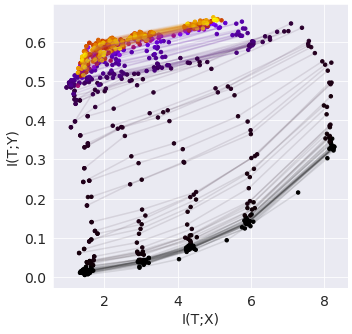}
        \end{subfigure}%
        \begin{subfigure}{0.32\linewidth}
            \includegraphics[width=\linewidth]{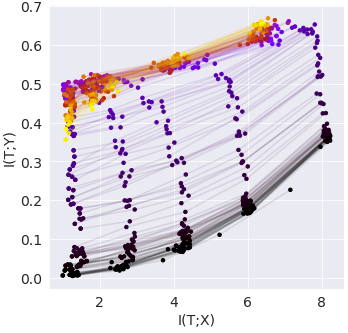}
        \end{subfigure}
        \begin{subfigure}{0.32\linewidth}
            \includegraphics[width=\linewidth]{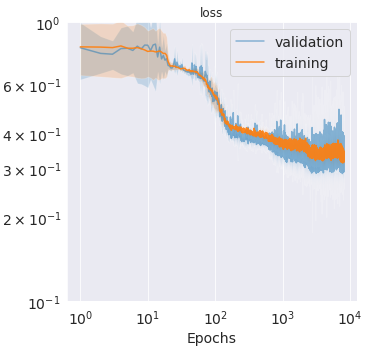}
        \end{subfigure}%
        \begin{subfigure}{0.32\linewidth}
            \includegraphics[width=\linewidth]{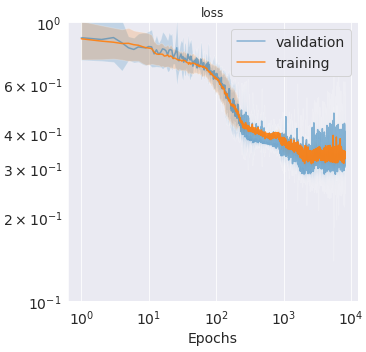}
        \end{subfigure}%
        \begin{subfigure}{0.32\linewidth}
            \includegraphics[width=\linewidth]{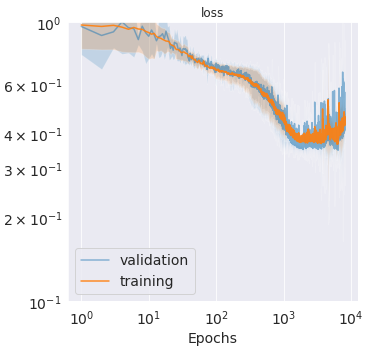}
        \end{subfigure}
        \begin{subfigure}{0.32\linewidth}
        \includegraphics[width=\linewidth]{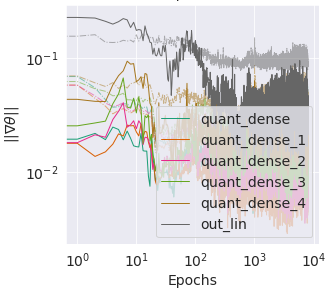}
        \caption{STE activation}
        \label{fig:ibbnn_spherical_ste}
    \end{subfigure}%
    \begin{subfigure}{0.32\linewidth}
        \includegraphics[width=\linewidth]{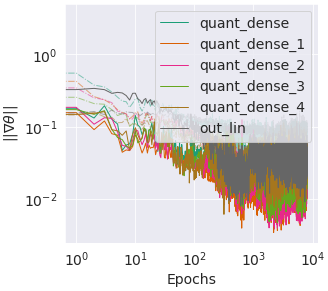}
        \caption{Approximate sign activation}
        \label{fig:ibbnn_spherical_approx}
    \end{subfigure}%
    \begin{subfigure}{0.32\linewidth}
        \includegraphics[width=\linewidth]{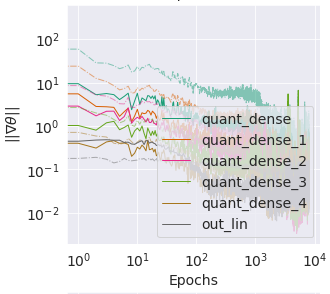}
        \caption{Swish sign activation}
        \label{fig:ibbnn_spherical_swish}
    \end{subfigure}
    \end{minipage}%
    \begin{minipage}{0.04\linewidth}
        \begin{subfigure}{\linewidth}
            \vspace*{-8.5cm}
            \includegraphics[width=\linewidth]{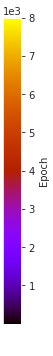}
        \end{subfigure}
    \end{minipage}
    \caption{Information plane behavior of different activations in BNNs on synthetic dataset. First row corresponds to the behavior on train set. Left most trace of dots in each figure corresponds to the output layer activation. Moving right, we get to each hidden layer starting from output layer. Second row shows evolution of loss for corresponding activations. Third row gives the evolution of norm of the gradients for corresponding activations. Solid lines correspond to the mean of observed metric; dim dashed line corresponds to the variance at each epoch.}
    \label{fig:ibbnn_spherical}
\end{figure}

Similar to \cite{shwartz2017opening, saxe2019information}, we use a network with $5$ hidden layers. Starting with $10$ neurons in the first hidden layer, the number of neurons in each layer is decreased by $2$ resulting in $10 - 8 - 6 - 4 - 2$ number of neurons in each layer. The final layer is a softmax with two outputs, one per each class. Batch size is chosen as $64$ and Adam optimizer with learning rate $0.0001$ is used for training. Hence $52$ weight update steps are applied to the model per epoch. The results are for $8000$ epochs of training. A binning estimator with $30$ bins per dimension is used for estimating MI. Even though $30$ bins are used, in the case of binary activations from intermediate layers, only two bins will be active, corresponding to $-1$ and $+1$ values. The information plane behavior for different activation functions along with the loss evolution and norm of the gradients during training is given in Fig. \ref{fig:ibbnn_spherical}.

The results of using STE activation function are provided in Fig. \ref{fig:ibbnn_spherical_ste}. We can see that the layers are initially stuck at some point for the first few epochs before it starts moving in the information plane. To facilitate the learning, latent weights that can take real values are used during the training phase. Following initialization, these values should be seen as the momentum of inertia \cite{helwegen2019latent} in rest and it may take a few epochs to break out of this, and then, the binary weights start flipping the sign. At this phase, we will see that the model starts to show some progress. This is also evident from the loss evolution where initial few epochs see a high variance in loss among the models, but almost no decrease in the mean loss. Following this, we see a loss decrease phase followed by a noisy saturation. This rapid phase of decreasing loss is also characterized by the rapid movement of MI points for each layer in the information plane with an increase in $I(T;Y)$. Comparing this to the information plane behavior in DNN (See results in \cite{shwartz2017opening,saxe2019information}. See Fig. \ref{fig:ibdnn_spherical} in Appendix for our reproduction), we can notice some interesting behavior. While DNNs first increase both $I(T;X)$ and $I(T;Y)$ followed by a separate representation compression (RC) phase where $I(T;X)$ decreases, BNNs start with a low value for $I(T;X)$ and does not show an explicit compression phase. Also, we can see that the compression phase is a slow process in DNN, and often happens once loss starts saturating. However, when we use methods like early stopping and saturation detection, practical models may never get to the compression phase. The compression phase of DNNs is seen as generalization \cite{shwartz2017opening} and this is not achievable unless models are trained well beyond loss saturation and are at the risk of overfitting (as seen from Fig. \ref{fig:ibdnn_spherical} in Appendix). BNNs are known to be immune to adversarial attacks to a considerable extent \cite{galloway2018attacking}. We hypothesize that this is because of the lack of the ability of the BNNs to increase $I(T;X)$. Because of the lack of expressivity of BNNs, arising from constrained weight and activation values, the model has to train by generalizing over the dataset rather than extracting features that may be specific for individual samples. As evident from Fig. \ref{fig:ibbnn_spherical}, hidden representations in BNNs carry very less information about the input itself, and during training they spend time on improving task-relevant mutual information $I(T;Y)$.

One of the main observations that support an ERM phase followed by a representation compression phase in DNNs is the behavior of gradients. Results in \cite{shwartz2017opening} show that during the initial phase of training of DNN, the variance in the gradients is low but becomes high as training progresses (See the last row of Fig. \ref{fig:ibdnn_spherical} in Appendix). \cite{shwartz2017opening} hypothesized that this behavior of high gradient variance in each epoch is similar to the noise and this facilitates the generalization in DNNs. In sharp contrast to these observations, we do not observe this  phase transition in BNNs (See the last row in Fig. \ref{fig:ibbnn_spherical}). Rather, we observe that both mean and variance of gradients remain at the same scale throughout the training process, yet BNNs provide good generalization. From the loss evolution in Fig. \ref{fig:ibbnn_spherical}, there is no sign of overfitting in BNN, while the same in Fig. \ref{fig:ibdnn_spherical} for DNNs show overfitting characteristics. This raises questions about characterizing the representation compression (RC) phase with high variance in gradients alone. There seems to be no explicit RC phase in BNNs and instead, generalization occurs simultaneously with label fitting.

\subsection{Results on MNIST dataset}
\begin{figure}[!t]
    \centering
    \begin{minipage}{0.94\linewidth}
        \begin{subfigure}{0.32\linewidth}
            \includegraphics[width=\linewidth]{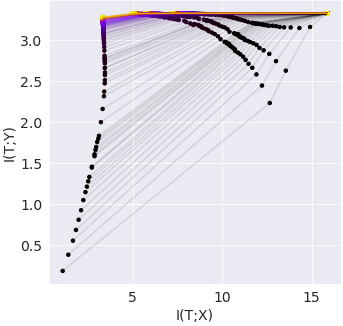}
        \end{subfigure}%
        \begin{subfigure}{0.32\linewidth}
            \includegraphics[width=\linewidth]{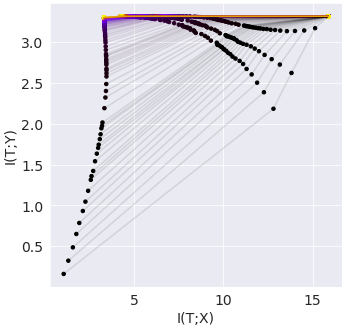}
        \end{subfigure}%
        \begin{subfigure}{0.32\linewidth}
            \includegraphics[width=\linewidth]{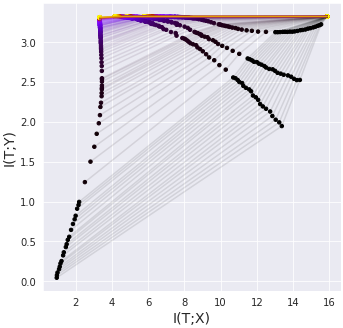}
        \end{subfigure}
        \begin{subfigure}{0.32\linewidth}
            \includegraphics[width=\linewidth]{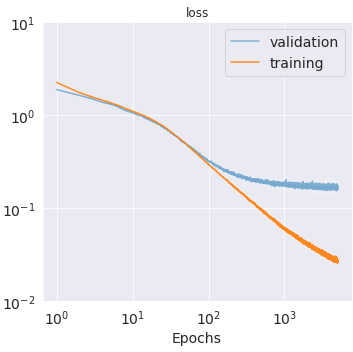}
        \end{subfigure}%
        \begin{subfigure}{0.32\linewidth}
            \includegraphics[width=\linewidth]{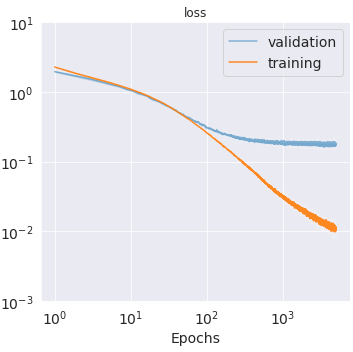}
        \end{subfigure}%
        \begin{subfigure}{0.32\linewidth}
            \includegraphics[width=\linewidth]{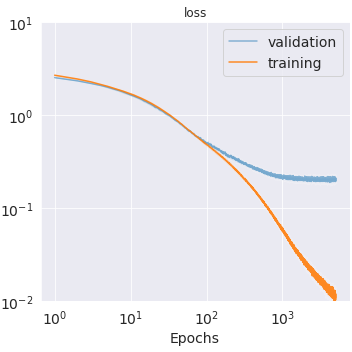}
        \end{subfigure}
        \begin{subfigure}{0.32\linewidth}
            \includegraphics[width=\linewidth]{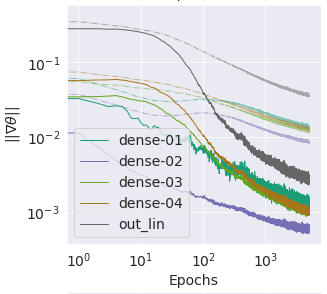}
            \caption{STE}
        \end{subfigure}%
        \begin{subfigure}{0.32\linewidth}
            \includegraphics[width=\linewidth]{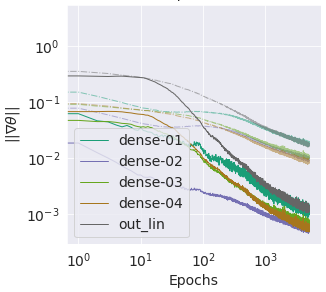}
            \caption{Approximate sign}
        \end{subfigure}%
        \begin{subfigure}{0.32\linewidth}
            \includegraphics[width=\linewidth]{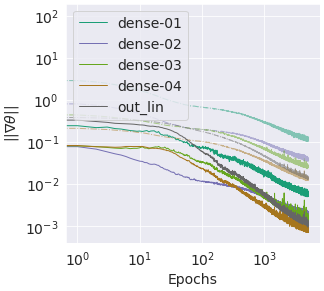}
            \caption{Swish sign}
        \end{subfigure}
    \end{minipage}%
    \begin{minipage}{0.04\linewidth}
        \begin{subfigure}{\linewidth}
            \vspace*{-8.5cm}
            \includegraphics[width=\linewidth]{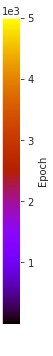}
        \end{subfigure}
    \end{minipage}
    \caption{Information plane behavior of different activations in BNNs with MNIST dataset. First row corresponds to the behavior on train set. Second row gives loss for corresponding activations. Third row gives the evolution of norm of gradients for corresponding activations. Solid lines correspond to the mean of observed metric; dim dashed line corresponds to the variance at each epoch. Results are averaged with observations from $5$ independent runs.}
    \label{fig:ibbnn_mnist}
\end{figure}

The results for training MNIST dataset classification task is provided in Fig. \ref{fig:ibbnn_mnist}. A fully connected network with $4$ hidden layers is used. Hidden nodes for each layer is set as $1024 - 20 - 20 - 20$, similar to \cite{saxe2019information}. The final layer has a softmax activation over $10$ output nodes. For training, a batch size of $128$ is used with Adam optimizer of learning rate $0.00001$. Hence, $469$ weight updates per epoch are applied during the training phase. Models are trained for $5000$ epochs and the presented results are averaged over $5$ independent runs. 

As in the case of the synthetic dataset, we can see that there is only a compression phase in BNNs training for all the three activation functions. Interestingly, the magnitude of compression in the case of MNIST is higher than that in the case of the synthetic dataset. This can be attributed to the low dimensional manifold of MNIST data. Even though there are $28 \times 28$ features for each sample, the data is heavily structured (image data) and most of the features are background pixels with no information that is relevant for prediction. Hence, each layer in the neural network can easily compress data while discarding irrelevant features. While DNNs try to capture most of the data during the initial training phase \cite{saxe2019information}, BNNs, because of its low expressivity, have to strive for compact representations from initial epochs. This is the reason for the high compression we see in the case of MNIST. For the case when activations can take real value, as in the case of output softmax layer, the compression is barely visible.

The evolution of gradient norm along with the loss evolution during training is given in the last row of Fig. \ref{fig:ibbnn_mnist}. Here also we can observe that the variance of the gradients is not increasing as in the case of DNNs (as reported by \cite{shwartz2017opening}). Rather, the variance of the gradient also decreases along with the mean of gradients. This observation reaffirms our initial understanding that high variance gradients in the diffusion phase are not required in BNNs for representation compression. The representation compression seems to happen simultaneously with training as we observed with the case of the synthetic dataset.

\subsection{Discussion}
Our analysis of the training dynamics of BNNs uncovered some interesting observations specific to BNNs. Previous works that analyzed DNNs \cite{shwartz2017opening} showed that training in DNNs happens as a two-phase procedure: initial ERM phase that minimizes the training loss followed by a RC phase characterized by chaotic gradients. This explicit RC phase is considered as the phase where generalization takes place in DNNs. Compared to the traditional DNNs, we showed that there is no explicit RC phase for BNNs; instead, the generalization should be assumed as happening concurrently during the training. We propose that this observation arises partly from the low representative power of BNNs due to the constrained weights and activations. 

As discussed, both activations and weights of BNNs are constrained to only two possible levels, $\{-1, +1\}$. Comparing this to the entire range of real values a traditional full precision DNN can take, it is immediately clear that BNNs are at the disadvantage of having very limited representational power. 

IB principle \cite{tishby2000information, tishby2015deep} formulates intermediate hidden layer activations in a neural network as a successive Markov chain. Because of the DPI \cite{tishby2015deep}, any information that is not included in an intermediate representation about input $X$ or target label $Y$ will not be available to the later layers. Since DNNs have the freedom to use the entire real number space for representing this information, each hidden layer in a DNN can capture most of the information from its input. However, this representational power can often cause problems if sample-specific information is propagated by hidden layers instead of the minimal sufficient statistics for prediction. This manifests as overfitting in trained models. Analysis of information plane behavior of DNNs \cite{shwartz2017opening} showed that initial epochs of training are spent in increasing both the $I(T;X)$ as well as $I(T;Y)$. The expressive power of DNNs can often help it in capturing more information about samples $X$ than that is necessary for target prediction. However, during the slow RC phase, this additional information is dropped as seen as a drop in $I(T;X)$ in DNNs.

The case of BNN is notably different. Because of the limited capacity to make hidden representations (due to binary weights and activations), BNNs do not have the luxury to pass a high amount of information about the input samples $X$. Even though this hinders the progress of training of BNNs initially, we observe that this comes with a great benefit. The hidden representations developed by BNNs tend to be parsimonious right from the beginning of the training; we see almost no reduction in $I(T;X)$ during the training. The limited expressive power of BNNs forces it to pass the most relevant information for target prediction at each layer. Further, we do not see an overfitting behavior in BNNs; this can possibly be attributed to the parsimonious hidden representations as confirmed by the information plane behavior. This could also answer how BNNs develop adversarial immunity during training itself \cite{galloway2018attacking}, without explicit procedures. Adversarial samples are usually developed by making small perturbations to the input features. Since the compressed representations of input by each hidden layer in BNNs are inherently limited in expressivity, those adversarial perturbations are usually neglected during the activation function at each hidden layer.

While BatchNorm is seen as an inseparable element for training BNNs, our experiments show that BatchNorm has some drawbacks. Our results, given in Appendix (Fig. \ref{fig:ibbnn_appendix_ste-sign}, Fig. \ref{fig:ibbnn_appendix_approx-sign} and Fig. \ref{fig:ibbnn_appendix_swish-sign} for BNNs with activations STE-sign, approx-sign, and swish-sign respectively) show layer-wise evolution of mutual information for BNNs in the synthetic dataset. It is clearly visible that the first BatchNorm layer losses information about target $Y$ (seen as a decrease in $I(T;Y)$ for BatchNorm layers). Because of the Markov structure of hidden representations, later layers will never be able to recover this information and hence suffer the damage in predictive power. Even though we did not observe this behavior in the MNIST dataset, it is clear that BatchNorm has some shortcomings. This also points to the need for developing better activity regularization methods for training BNNs \cite{ding2019regularizing}.

    \section{Concluding Remarks}
This paper presented an information-theoretic viewpoint of training dynamics of Binary Neural Networks. As opposed to DNNs, our experiments showed that BNNs have a notably different information propagation behavior. While DNNs have two separate phases for risk minimization and generalization during training, BNNs seem to have a concurrent one. Further, the training dynamics of BNNs show that the low expressive power of BNNs forces it to make generalized parsimonious hidden representations that are necessary for the task. While this work sheds light on the training dynamics of BNNs with the help of the Information Bottleneck principle, the observations could assist in designing better optimization techniques for training BNNs. Using IB objective to train BNNs merit further investigation and can be an interesting avenue for future research.

    \small
    \bibliographystyle{unsrt}
    \bibliography{99_library.bib}
    
    \appendix
\section{Learning dynamics of DNN in information plane}\label{App:act}

\begin{figure}[!ht]
    \centering
    \begin{subfigure}{0.32\linewidth}
        \resizebox{\linewidth}{!}{
        \begin{tikzpicture}
            \begin{axis}[
                width=8cm,height=6cm,
                ylabel=$g(x)$,xlabel=$x$,
                ymin=-1.25,ymax=1.25,xmin=-3,xmax=3,
                grid=major,
                legend pos=south east,
                legend cell align={left},
                legend style={fill opacity=0.6, draw opacity=1.0, text opacity=1.0, font=\small}
            ]
            \addplot[blue,smooth,domain=-3:3] {tanh(x)};
            \addlegendentry{Forward pass}
            \addplot[red,smooth,domain=-3:3] {1/(cosh(x))^2};
            \addlegendentry{Backward pass}
            
            \end{axis}
        \end{tikzpicture}
        }
        \caption{tanh}
        \label{fig:tanh}
    \end{subfigure}
    \begin{subfigure}{0.32\linewidth}
        \resizebox{\linewidth}{!}{
        \begin{tikzpicture}
            \begin{axis}[
                width=8cm,height=6cm,
                ylabel=$g(x)$,xlabel=$x$,
                ymin=-1.25,ymax=1.25,xmin=-3,xmax=3,
                grid=major,
                legend pos=south east,
                legend cell align={left},
                legend style={fill opacity=0.6, draw opacity=1.0, text opacity=1.0, font=\small}
            ]
            \addplot[blue,smooth,domain=-3:-1] {-1};
            \addlegendentry{Forward pass}
            \addplot[red,smooth,domain=-3:-1] {0};
            \addlegendentry{Backward pass}
            \addplot[blue,smooth,domain=-1:+1] {x};
            \addplot[blue,smooth,domain=+1:+3] {+1};
            \addplot[red,smooth,domain=-1:+1] {1};
            \addplot[red,smooth,domain=+1:+3] {0};

            \end{axis}
        \end{tikzpicture}
        }
        \caption{hard-tanh}
        \label{fig:hard-tanh}
    \end{subfigure}
    \begin{subfigure}{0.32\linewidth}
        \resizebox{\linewidth}{!}{
        \begin{tikzpicture}
            \begin{axis}[
                width=8cm,height=6cm,
                ylabel=$g(x)$,xlabel=$x$,
                ymin=-5.25,ymax=5.25,xmin=-3,xmax=3,
                grid=major,
                legend pos=south east,
                legend cell align={left},
                legend style={fill opacity=0.6, draw opacity=1.0, text opacity=1.0, font=\small}
            ]
            \addplot[blue,smooth,domain=-3:+3] {2*(1./(1.+e^(-5*x)))*(1. +  5*x*(1. - 1./(1.+e^(-5.*x)))) - 1.};
            \addlegendentry{Forward pass}
            \addplot[red,smooth,domain=-3:+3] {(5 * (2 - 5*x*tanh(5*x/2)))/(1 + cosh(5*x))};
            \addlegendentry{Backward pass}  
            \end{axis}
        \end{tikzpicture}
        }
        \caption{sign-swish  \cite{darabiregularized} $(\beta = 5.0)$}
        \label{fig:signswish}
    \end{subfigure}
    \caption{Different activation functions for DNN models. Blue lines indicate the activation function $g(\cdot)$ in forward pass during training and inference. Red lines show the functions used as differential $g'(\cdot)$ used during backpropagation.}
    \label{fig:dnn_activations}
\end{figure}
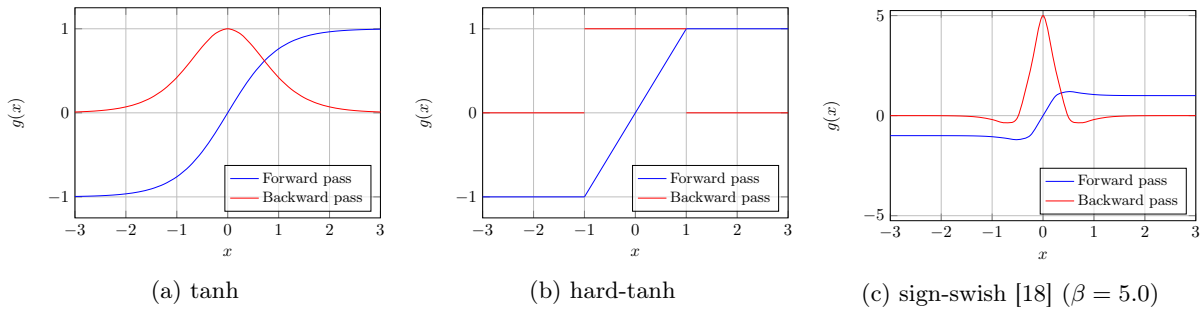

\begin{figure}[!ht]
    \centering
    \begin{minipage}{0.94\linewidth}
        \begin{subfigure}{0.32\linewidth}
            \includegraphics[width=\linewidth]{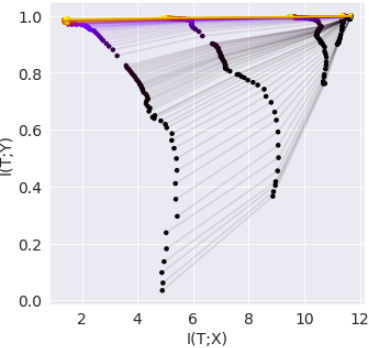}
        \end{subfigure}%
        \begin{subfigure}{0.32\linewidth}
            \includegraphics[width=\linewidth]{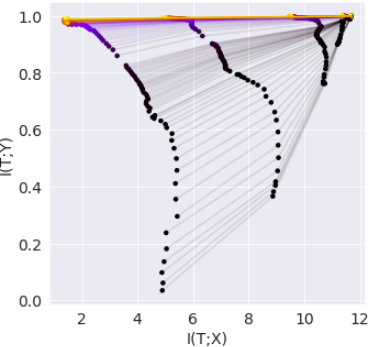}
        \end{subfigure}%
        \begin{subfigure}{0.32\linewidth}
            \includegraphics[width=\linewidth]{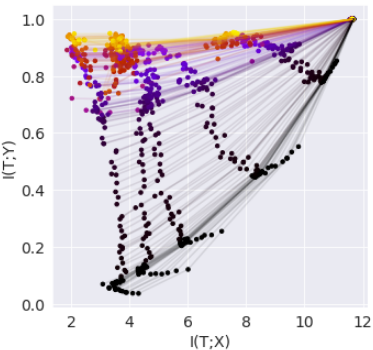}
        \end{subfigure}
        \begin{subfigure}{0.32\linewidth}
            \includegraphics[width=\linewidth]{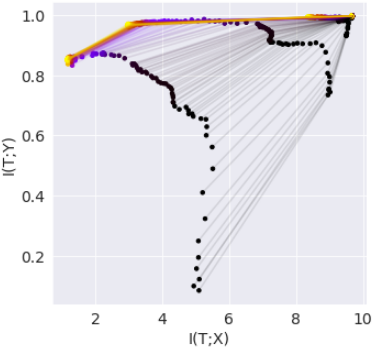}
        \end{subfigure}%
        \begin{subfigure}{0.32\linewidth}
            \includegraphics[width=\linewidth]{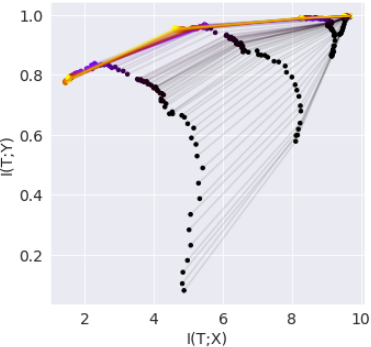}
        \end{subfigure}%
        \begin{subfigure}{0.32\linewidth}
            \includegraphics[width=\linewidth]{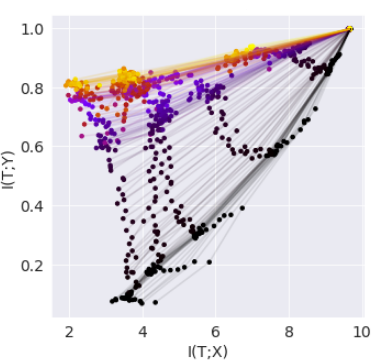}
        \end{subfigure}
        \begin{subfigure}{0.32\linewidth}
            \includegraphics[width=\linewidth]{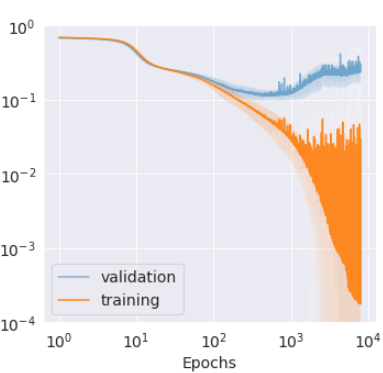}
        \end{subfigure}%
        \begin{subfigure}{0.32\linewidth}
            \includegraphics[width=\linewidth]{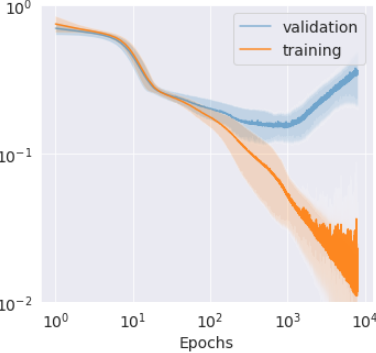}  
        \end{subfigure}%
        \begin{subfigure}{0.32\linewidth}
            \includegraphics[width=\linewidth]{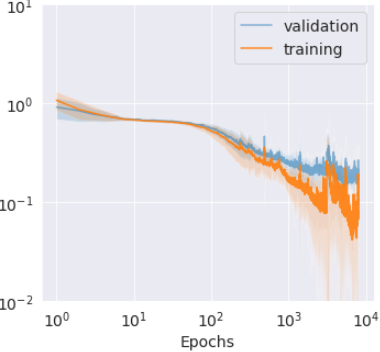}
        \end{subfigure}
        \begin{subfigure}{0.32\linewidth}
            \includegraphics[width=\linewidth]{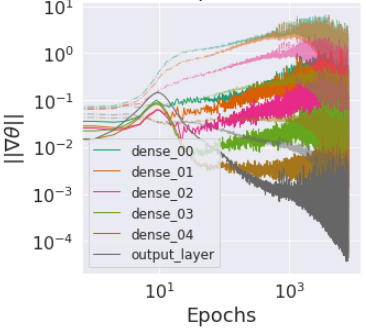}
            \caption{Tanh}
            \label{fig:ibdnn_spherical_tanh}
        \end{subfigure}%
        \begin{subfigure}{0.32\linewidth}
            \includegraphics[width=\linewidth]{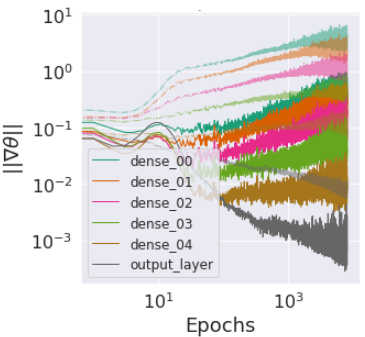}
            \caption{Hard tanh}
            \label{fig:ibdnn_spherical_htanh}
        \end{subfigure}%
        \begin{subfigure}{0.32\linewidth}
            \includegraphics[width=\linewidth]{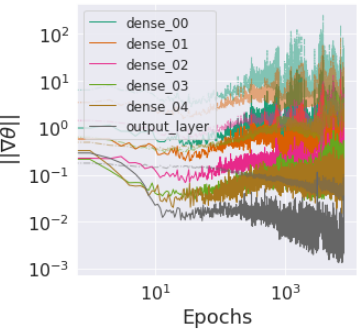}
            \caption{Sign swish}
            \label{fig:ibdnn_spherical_signswish}
        \end{subfigure}
    \end{minipage}%
    \begin{minipage}{0.06\linewidth}
        \begin{subfigure}{\linewidth}
            \vspace*{-8cm}
            \includegraphics[width=\linewidth]{figs/cropped/sidebar-epochs-8k.png}
        \end{subfigure}
    \end{minipage}
    \caption{Information plane behavior of different activations in DNNs with the synthetic dataset. First row corresponds to the behavior on train set. Second row corresponds to that of test set. X and Y axis denote the mutual information of the compressed intermediate layer representations with input X and output Y respectively. We show the changes in $I(T;X)$ and $I(T;Y)$ over training duration for all intermediate layers; left to right representing last layer to first layer. In the third row, we provide the loss evolution for train and validation set during the training process. The last row gives the evolution of gradients over time. All results are averaged over five independent runs. The shaded region in loss evolution shows the $1-\sigma$ confidence level.}
    \label{fig:ibdnn_spherical}
\end{figure}

The two activations hard-tanh and sign-swish are double saturating non-linearities like tanh as shown in Fig. \ref{fig:dnn_activations}. Both hard-tanh and sign-swish are very close to binary activation, but unlike binary activation, these non-linearities are continuous.
In this experiment, we study DNN with three activation functions $\tanh$, hard-tanh and sign-swish. The activation hard-tanh is given by 
\begin{align}
    g(x) = \begin{cases}
        -1 \qquad ; x \leq -1 \\
        x \quad \quad ; -1 \leq x \leq +1, \\
        +1 \qquad ; x \geq 1.
    \end{cases}
\end{align}
The sign-swish activation is given by 
\begin{equation}
    g(x) = 2\sigma(\beta x) \left(1 + \beta x\left(1 - \sigma(\beta x)\right)\right) - 1.
\end{equation}
where $\sigma$ is the sigmoid function, $\beta$ is a tunable parameter. In Fig. \ref{fig:ibdnn_spherical} we provide results on the synthetic dataset with these activation functions.

\section{Additional results with BNN}
\begin{figure*}[!t]
    \centering
    \begin{minipage}{0.94\linewidth}
    \begin{subfigure}[b]{0.48\textwidth}
        \includegraphics[width=0.48\textwidth]{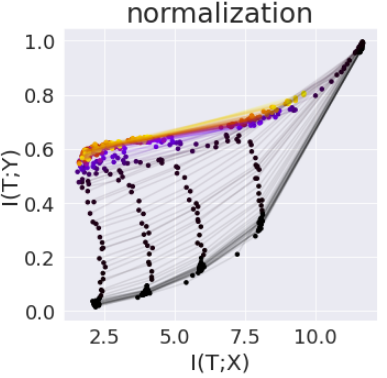} 
        \includegraphics[width=.49\textwidth]{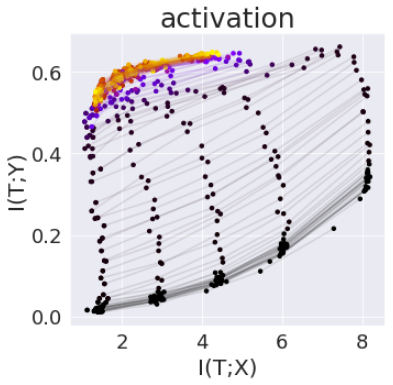}
        \caption{Training}
    \end{subfigure} 
    \begin{subfigure}[b]{0.48\textwidth}
        \includegraphics[width=0.48\textwidth]{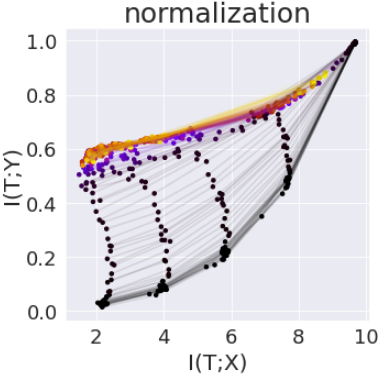} 
        \includegraphics[width=0.48\textwidth]{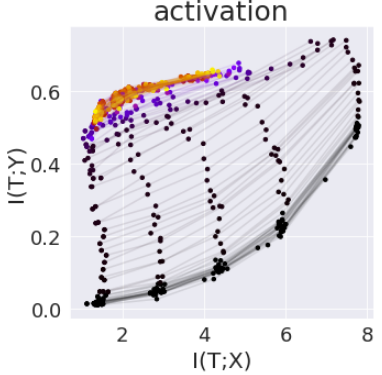}
        \caption{Test}
    \end{subfigure}
    \begin{subfigure}{1.0\textwidth}
        \includegraphics[width=\linewidth]{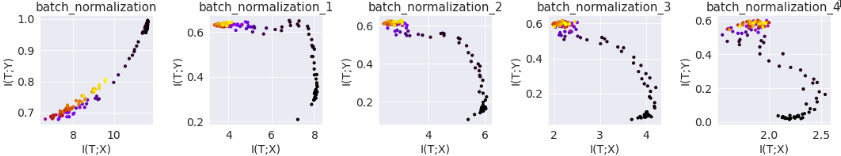}
    \end{subfigure}
    \begin{subfigure}{1.0\textwidth}
        \includegraphics[width=\linewidth]{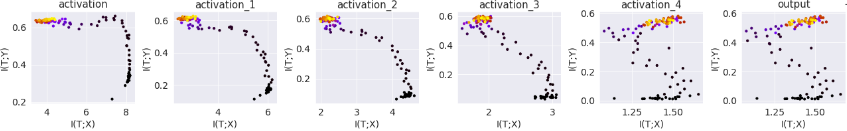}
        \caption{Layerwise dynamics during training}
    \end{subfigure}
    \begin{subfigure}{1.0\textwidth}
        \includegraphics[width=\linewidth]{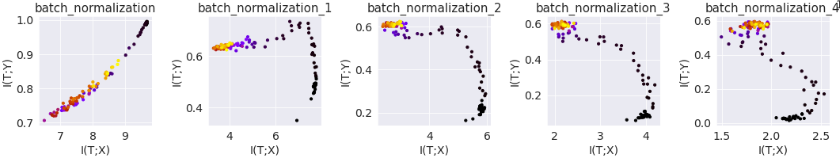}
    \end{subfigure}
    \begin{subfigure}{1.0\textwidth}
        \includegraphics[width=\linewidth]{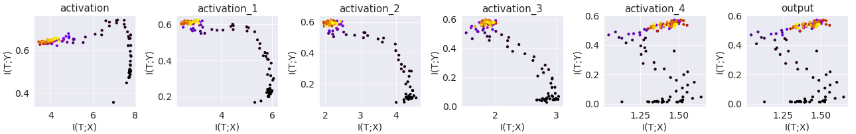}
        \caption{Layerwise dynamics during test}
    \end{subfigure}
    \begin{subfigure}{0.35\textwidth}
        \includegraphics[width=\linewidth]{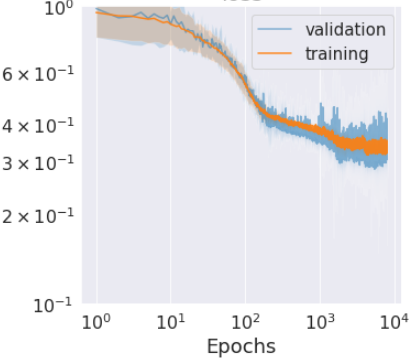}
        \caption{Loss}
    \end{subfigure}%
    \begin{subfigure}{0.30\textwidth}
        \includegraphics[width=\linewidth]{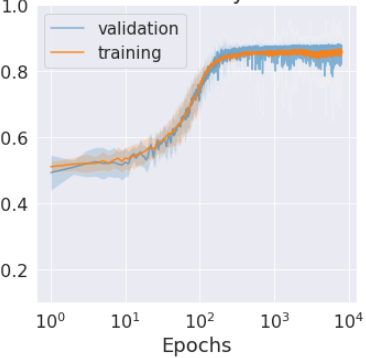}
        \caption{Accuracy}
    \end{subfigure}%
    \begin{subfigure}{0.35\textwidth}
        \includegraphics[width=\linewidth]{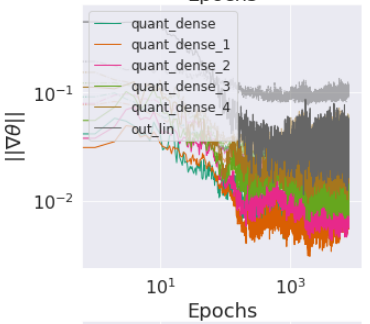}
        \caption{Gradient evolution}
    \end{subfigure}
    \end{minipage}%
    \begin{minipage}{0.06\linewidth}
        \begin{subfigure}{\linewidth}
            \vspace*{-7cm}
            \includegraphics[width=\linewidth]{figs/cropped/sidebar-epochs-8k.png}
        \end{subfigure}
    \end{minipage}
    \caption{Learning dynamics of BNN with activation ste-sign during training using synthetic dataset}
    \label{fig:ibbnn_appendix_ste-sign}
\end{figure*}

\begin{figure*}[!t]
    \centering
    \begin{minipage}{0.94\linewidth}
    \begin{subfigure}[b]{0.48\textwidth}
        \includegraphics[width=0.48\textwidth]{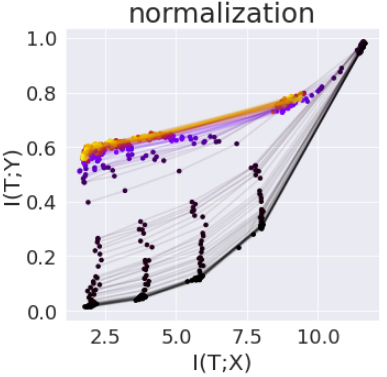} 
        \includegraphics[width=0.48\textwidth]{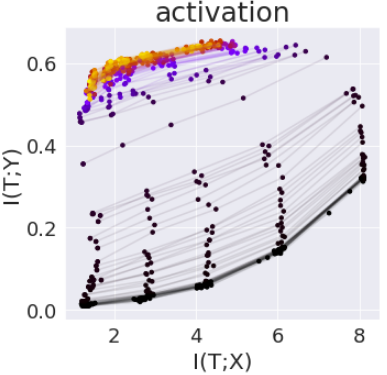}
        \caption{Training}
    \end{subfigure} 
    \begin{subfigure}[b]{0.48\textwidth}
        \includegraphics[width=0.48\textwidth]{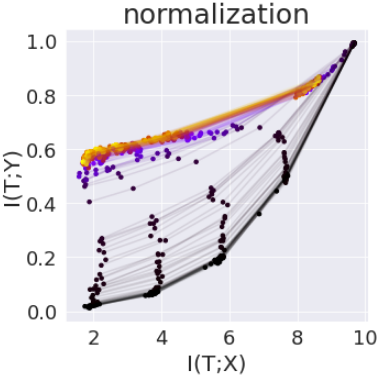} 
        \includegraphics[width=0.48\textwidth]{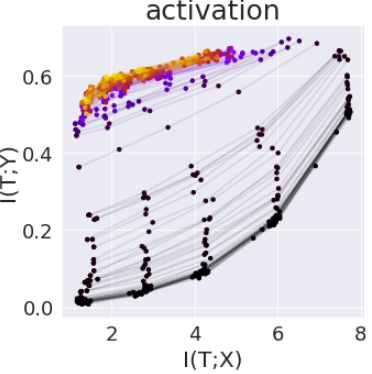}
        \caption{Test}
    \end{subfigure}
    \begin{subfigure}{1.0\textwidth}
        \includegraphics[width=\linewidth]{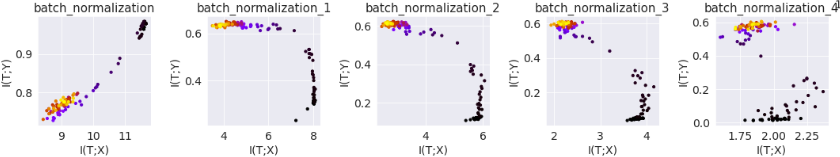}
    \end{subfigure}
    \begin{subfigure}{1.0\textwidth}
        \includegraphics[width=\linewidth]{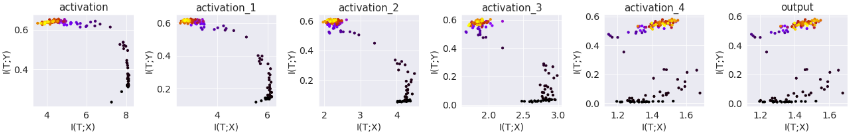}
        \caption{Layerwise dynamics during training}
    \end{subfigure}
    \begin{subfigure}{1.0\textwidth}
        \includegraphics[width=\linewidth]{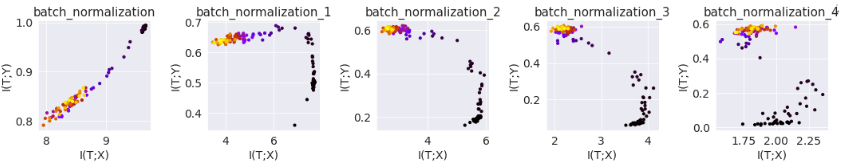}
    \end{subfigure}
    \begin{subfigure}{1.0\textwidth}
        \includegraphics[width=\linewidth]{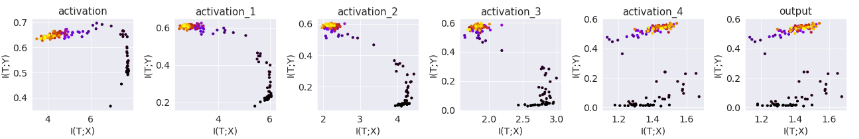}
        \caption{Layerwise dynamics during test}
    \end{subfigure}
    \begin{subfigure}{0.35\textwidth}
        \includegraphics[width=\linewidth]{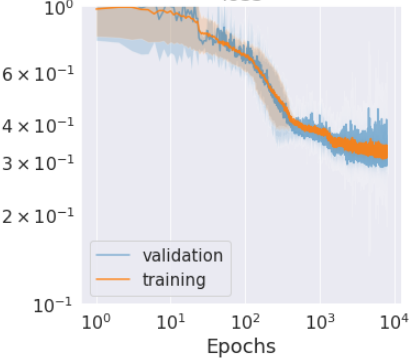}
        \caption{Loss}
    \end{subfigure}%
    \begin{subfigure}{0.30\textwidth}
        \includegraphics[width=\linewidth]{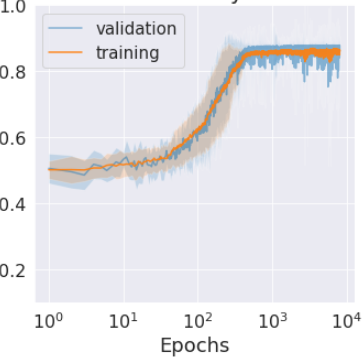}
        \caption{Accuracy}
    \end{subfigure}%
    \begin{subfigure}{0.35\textwidth}
        \includegraphics[width=\linewidth]{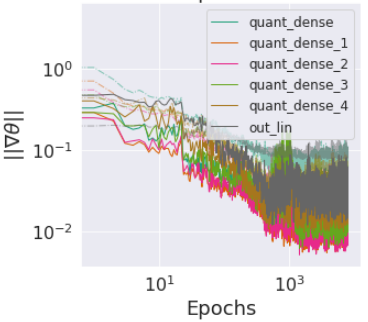}
        \caption{Gradient evolution}
    \end{subfigure}
    \end{minipage}%
    \begin{minipage}{0.06\linewidth}
        \begin{subfigure}{\linewidth}
            \vspace*{-7cm}
            \includegraphics[width=\linewidth]{figs/cropped/sidebar-epochs-8k.png}
        \end{subfigure}
    \end{minipage}
    \caption{Learning dynamics of BNN with activation approx-sign during training using synthetic dataset}
    \label{fig:ibbnn_appendix_approx-sign}
\end{figure*}

\begin{figure*}[!t]
    \centering
    \begin{minipage}{0.94\linewidth}
    \begin{subfigure}[b]{0.48\textwidth}
        \includegraphics[width=0.48\textwidth]{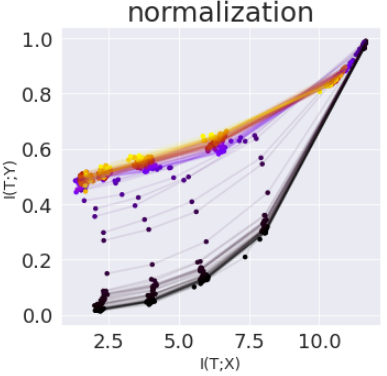} 
        \includegraphics[width=0.48\textwidth]{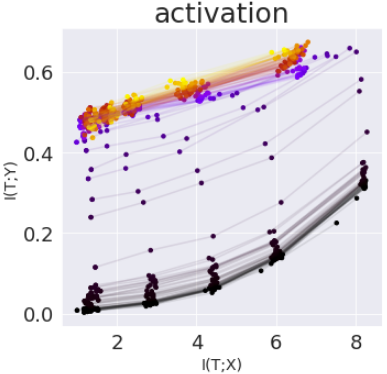}
        \caption{Training}
    \end{subfigure} 
    \begin{subfigure}[b]{0.48\textwidth}
        \includegraphics[width=0.48\textwidth]{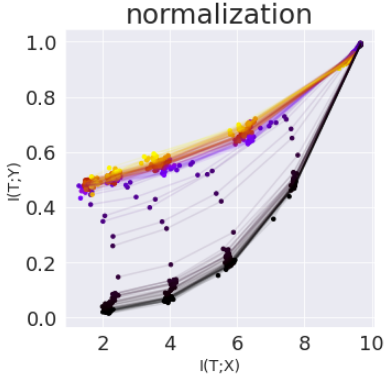} 
        \includegraphics[width=0.48\textwidth]{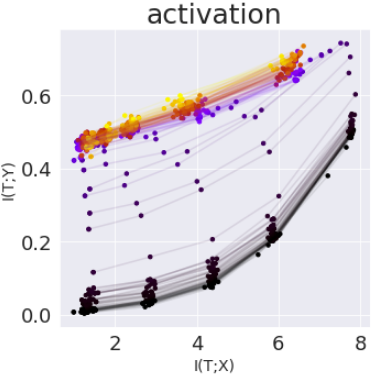}
        \caption{Test}
    \end{subfigure}
    \begin{subfigure}{1.0\textwidth}
        \includegraphics[width=\linewidth]{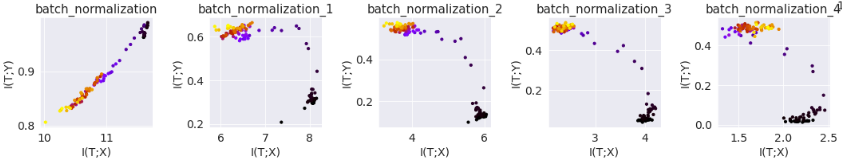}
    \end{subfigure}
    \begin{subfigure}{1.0\textwidth}
        \includegraphics[width=\linewidth]{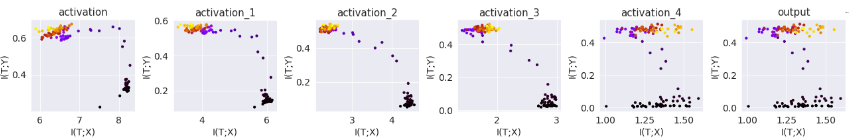}
        \caption{Layerwise dynamics during training}
    \end{subfigure}
    \begin{subfigure}{1.0\textwidth}
        \includegraphics[width=\linewidth]{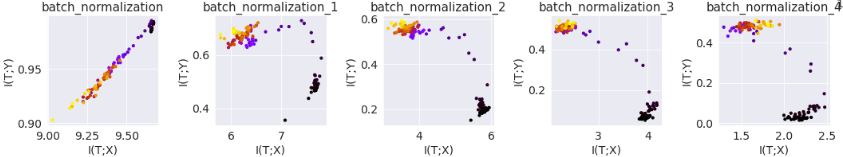}
    \end{subfigure}
    \begin{subfigure}{1.0\textwidth}
        \includegraphics[width=\linewidth]{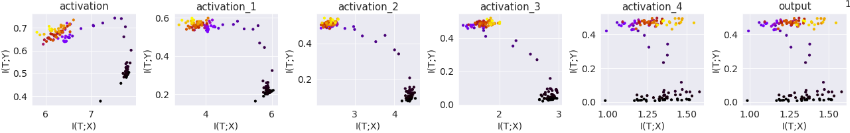}
        \caption{Layerwise dynamics during test}
    \end{subfigure}
    \begin{subfigure}{0.35\textwidth}
        \includegraphics[width=\linewidth]{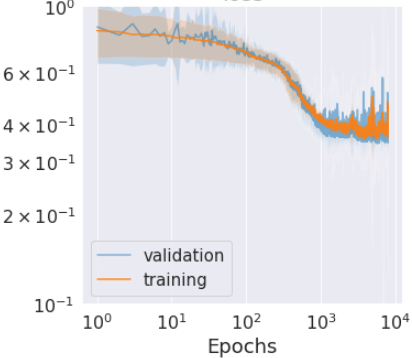}
        \caption{Loss}
    \end{subfigure}%
    \begin{subfigure}{0.30\textwidth}
        \includegraphics[width=\linewidth]{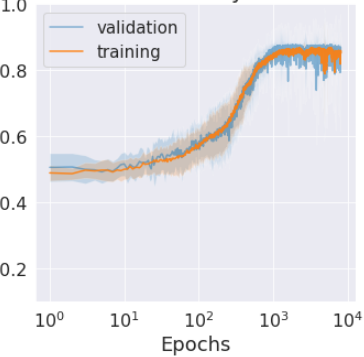}
        \caption{Accuracy}
    \end{subfigure}%
    \begin{subfigure}{0.35\textwidth}
        \includegraphics[width=\linewidth]{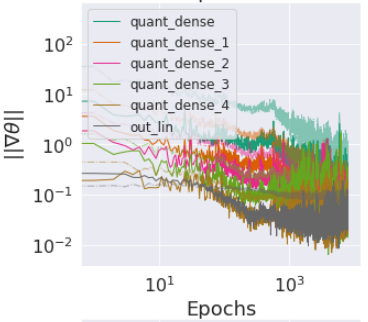}
        \caption{Gradient evolution}
    \end{subfigure}
    \end{minipage}%
    \begin{minipage}{0.06\linewidth}
        \begin{subfigure}{\linewidth}
            \vspace*{-7cm}
            \includegraphics[width=\linewidth]{figs/cropped/sidebar-epochs-8k.png}
        \end{subfigure}
    \end{minipage}
    \caption{Learning dynamics of BNN with activation swish-sign during training using synthetic dataset}
    \label{fig:ibbnn_appendix_swish-sign}
\end{figure*}

In this section we provide detailed results of training dynamics of BNN with three activations ste-sign, approx-sign and swish-sign.
The results include the layer-wise learning dynamics of BNN, the evolution of gradients, loss and accuracy plots and are presented in Fig. \ref{fig:ibbnn_appendix_ste-sign}, Fig. \ref{fig:ibbnn_appendix_approx-sign} and Fig. \ref{fig:ibbnn_appendix_swish-sign} for activations ste-sign, approx-sign and swish-sign respectively. Closely observing the learning dynamics of BNN layer-wise, as discussed in the main paper, we don't see the presence of two phases (ERM and RC phases) separately in case of BNN. The first normalization layer decreases $I(T;Y)$ for all of the three activation functions. This suggests exploration of better normalization techniques. The plots for training and validation loss in case of BNNs exhibit no overfitting, keeping the accuracy for validation close to training. There is no separate RC phase while training BNN where only generalization takes place. The variance in gradients is of the same scale as the mean of the gradients throughout the training. Thus supporting the fact that generalization occurs concurrently with the risk minimization.

\section{Results with Tic-Tac-Toe dataset}

\begin{figure*}[!t]
    \centering
    \begin{minipage}{0.94\linewidth}
    \begin{subfigure}[b]{.5\linewidth}
        \includegraphics[width=.45\textwidth]{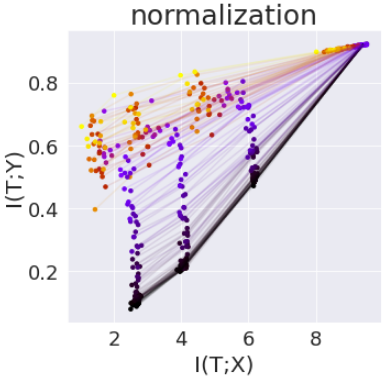}
        \includegraphics[width=.45\textwidth]{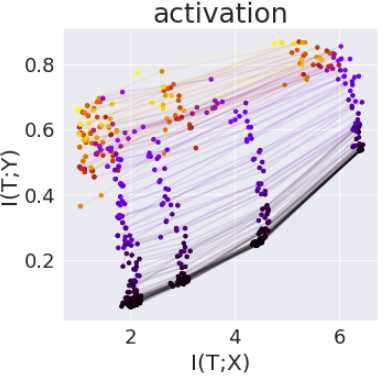}
        \caption{Training}
    \end{subfigure} 
    \begin{subfigure}[b]{.5\textwidth}
        \includegraphics[width=.45\textwidth]{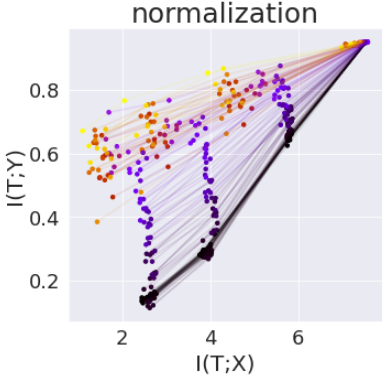}
        \includegraphics[width=.45\textwidth]{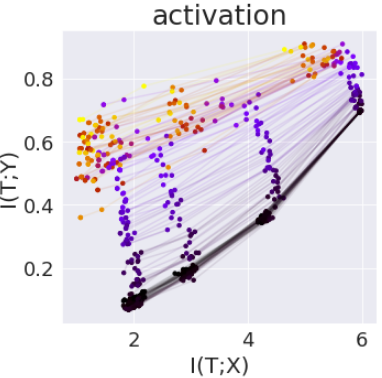}
        \caption{Test}
    \end{subfigure}
    \begin{subfigure}{0.95\textwidth}
        \includegraphics[width=\linewidth]{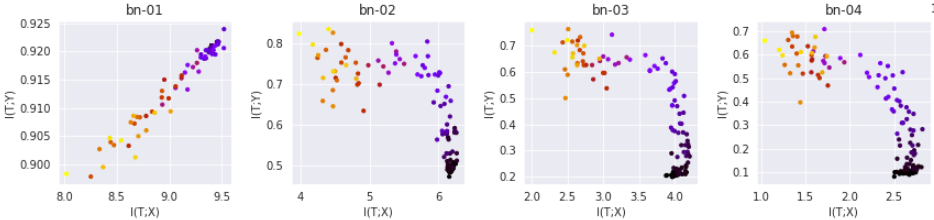}
    \end{subfigure}
    \begin{subfigure}{0.95\textwidth}
        \includegraphics[width=\linewidth]{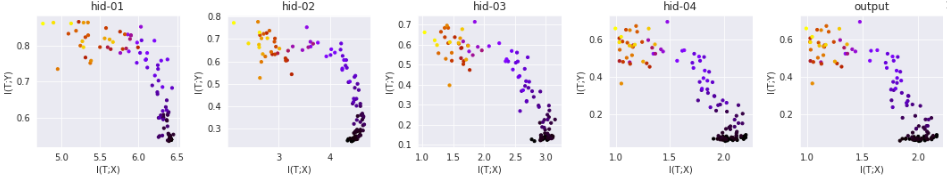}
        \caption{Layerwise dynamics during training}
    \end{subfigure}
    \begin{subfigure}{0.95\textwidth}
        \includegraphics[width=\linewidth]{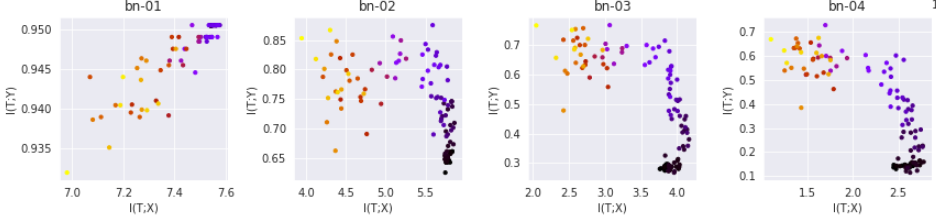}
    \end{subfigure}
    \begin{subfigure}{0.95\textwidth}
        \includegraphics[width=\linewidth]{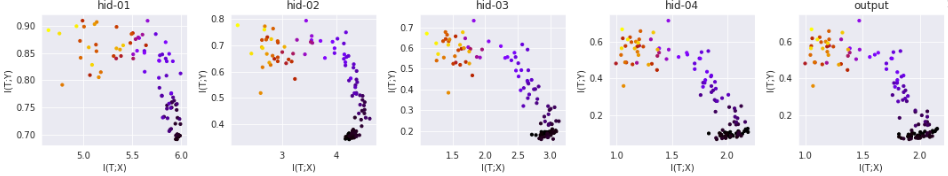}
        \caption{Layerwise dynamics during test}
    \end{subfigure}
    \begin{subfigure}{0.31\textwidth}
        \includegraphics[width=\linewidth]{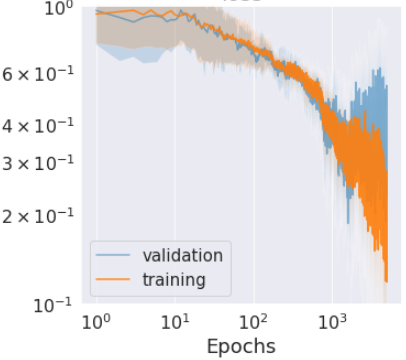}
        \caption{Loss}
    \end{subfigure}%
    \begin{subfigure}{0.29\textwidth}
        \includegraphics[width=\linewidth]{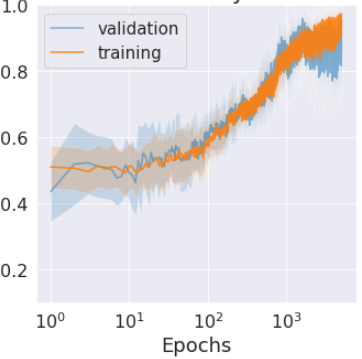}
        \caption{Accuracy}
    \end{subfigure}%
    \begin{subfigure}{0.31\textwidth}
        \includegraphics[width=\linewidth]{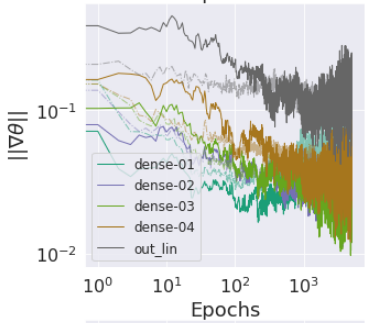}
        \caption{Gradient evolution}
    \end{subfigure}
    \end{minipage}%
    \begin{minipage}{0.06\linewidth}
        \begin{subfigure}{\linewidth}
            \vspace*{-7cm}
            \includegraphics[width=\linewidth]{figs/cropped/sidebar-epochs-8k.png}
        \end{subfigure}
    \end{minipage}
    \caption{Learning dynamics of BNN with activation ste-sign during training using Tic-Tac-Toe endgame dataset}
    \label{fig:ttt-ibbnn_appendix_ste-sign}
\end{figure*}

To support our observations, we provide the information plane behavior of the BNN with ste-sign activation with a popular Tic-Tac-Toe endgame dataset \cite{Dua:2019} in Fig. \ref{fig:ttt-ibbnn_appendix_ste-sign}. 

The dataset encodes the possible Tic-Tac-Toe board configurations at the end of the game where "x" is assumed to be played first. Each sample has $9$ features corresponding to $9$ board position of the game. The $9$ features in the input are board positions like top-left-square, top-middle-square, top-right-square, middle-left square and so on. The target is to get win for "x" and this happens only if "x" gets three consecutive places on the board. Therefore the data has two labels indicating win or lose for "x".  For example "x, x, x, x, o, b, o, o, b" is a win for "x" but "o, x, x, o, b, b, o, x, b" is not. The dataset has $958$ samples out of which $766$ samples are used to train the model and the rest is for validation. For our experiment we encode "x" as $+1$, "o" as $-1$ and "b" as 0. The labels win and lose are encoded as $+1$ and $0$ respectively.

Even with this dataset, the findings are similar to that of the synthetic dataset. Thus supporting our observations on learning dynamics of BNNs that BNNs do not have ERM and RC phase separately. Rather the compression happens concurrently with the label fitting.

\section{Test for generalization in BNN}
In supervised learning, generalization error is a measure of how accurately an algorithm can predict the target on a previously unseen data. In a cross-validation test, the whole dataset is partitioned into two for training and test where the former is used to train a network and then tested with the latter which was not seen before by the model. A network with a higher degree of freedom, capable of learning a more complex function than the required one, tries to overfit the noise present in the input of the training data instead of learning the underlying relationship. This usually results in the model overfitting the train samples and gives raise to generalization errors. However, this can be easily detected as the difference in loss between the train and test datasets. 

\begin{figure*}[!t]
    \centering
    \begin{subfigure}{0.33\textwidth}
        \includegraphics[width=\linewidth]{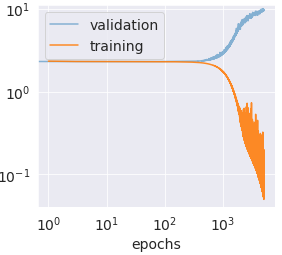}
    \end{subfigure}%
    \begin{subfigure}{0.33\textwidth}
        \includegraphics[width=\linewidth]{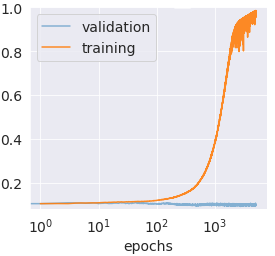}
    \end{subfigure}%
    \begin{subfigure}{0.33\textwidth}
        \includegraphics[width=\linewidth]{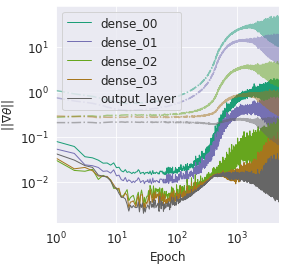}
    \end{subfigure}
    \begin{subfigure}{0.33\textwidth}
        \includegraphics[width=\linewidth]{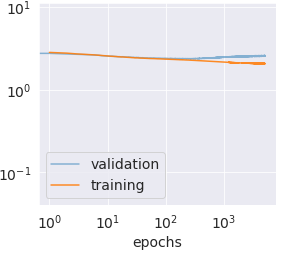}
        \caption{Loss}
        \label{fig:MNIST_randomlabel_metrics_loss}
    \end{subfigure}%
    \begin{subfigure}{0.33\textwidth}
        \includegraphics[width=\linewidth]{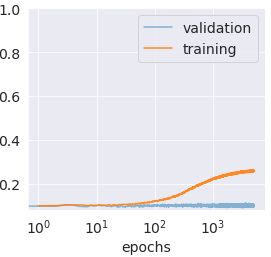}
        \caption{Accuracy}
        \label{fig:MNIST_randomlabel_metrics_accuracy}
    \end{subfigure}%
    \begin{subfigure}{0.33\textwidth}
        \includegraphics[width=\linewidth]{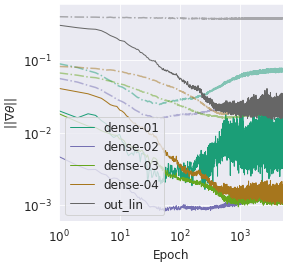}
        \caption{Gradient evolution} \label{fig:MNIST_randomlabel_metrics_grad_evo}
    \end{subfigure}
    \caption{Comparison of loss, accuracy and gradient evolution between DNN (top row) and BNN (bottom row) when trained with random labeled MNIST data. The loss and accuracy plots show clear presence of overfitting in DNN but not in BNN. Even though the DNN does not generalize, the network still have a noisy gradient phase later in the training as seen from gradient evolution graphs. In DNN, the variance of the gradient is high because each batch of samples will try to overfit the loss for that batch only, which will lead to gradients with random directions. This manifests as high variance in the norm of gradients for each epoch. However the BNNs do not overfit as seen from (a) and hence the variance between the gradients of each batch will be low.}
    \label{fig:MNIST_randomlabel_metrics}
\end{figure*}

\begin{figure*}[!t]
    \centering
    \begin{minipage}{0.94\linewidth}
    \begin{subfigure}{1.0\textwidth}
        \includegraphics[width=\linewidth]{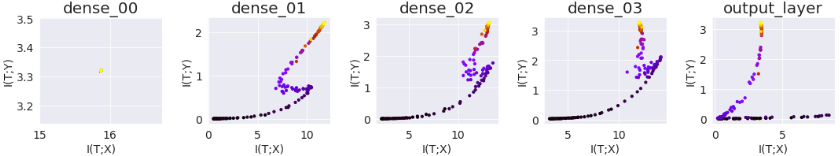}
    \end{subfigure}
    \begin{subfigure}{1.0\textwidth}
        \includegraphics[width=\linewidth]{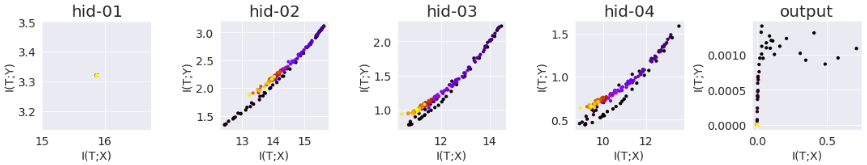}
        \caption{Learning dynamics on train set, top row presents learning dynamics of DNN and the bottom row is the same of BNN} \label{fig:MNIST_randomlabel_dynamics_train_DNN}
    \end{subfigure}
    \begin{subfigure}{1.0\textwidth}
        \includegraphics[width=\linewidth]{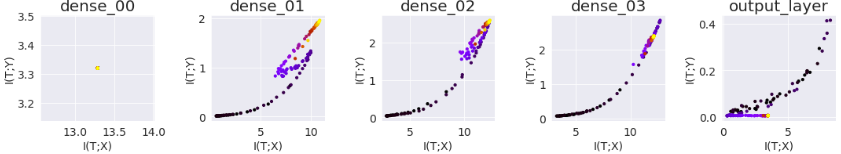}
    \end{subfigure}
    \begin{subfigure}{1.0\textwidth}
        \includegraphics[width=\linewidth]{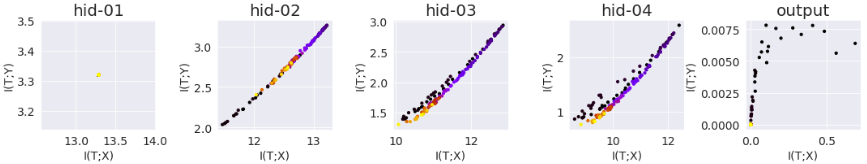}
        \caption{Learning dynamics on test set, top row presents learning dynamics of DNN and the bottom row is the same of BNN}
        \label{fig:MNIST_randomlabel_dynamics_test}
    \end{subfigure}
    \end{minipage}%
    \begin{minipage}{0.06\linewidth}
        \begin{subfigure}{\linewidth}
            \vspace*{0cm}
            \includegraphics[width=\linewidth]{figs/cropped/sidebar-epochs-8k.png}
        \end{subfigure}
    \end{minipage}
    \caption{A comparison of layerwise learning dynamics between DNN and BNN with activation tanh and ste-sign respectively to show how good a DNN or BNN model gets trained with random labels of MNIST dataset}
    \label{fig:MNIST_randomlabel_dynamics}
\end{figure*}

In order to study the generalization capabilities of DNNs and BNNs, we train both the models with random data. For this study, we took standard MNIST data and shuffled the labels. Ideally, in this case, the model should not be able to learn any useful concepts, and the accuracy should stay around $0.10$ on the train set if the model does not overfit the train samples. We compare how DNN and BNN behave differently while training on shuffled labels and provide results of a single trained model for each of the cases. We provide loss, accuracy, and gradient evolution in Fig. \ref{fig:MNIST_randomlabel_metrics} and layerwise learning dynamics during training and test in Fig. \ref{fig:MNIST_randomlabel_dynamics}. 

With a random labeled MNIST, ideally, a model should learn nothing. But in spite of random labels, the accuracy for DNN increases whereas BNN's accuracy does not increase as seen in Fig. \ref{fig:MNIST_randomlabel_metrics_accuracy}. Looking at the information plane behavior of DNN in Fig. \ref{fig:MNIST_randomlabel_dynamics}, the last layer along with all intermediate layers of DNN show an increase in $I(T;Y)$, indicating that the DNN can be trained to capture even sample-specific information to predict target correctly while training with random labels. An increase in $I(T;X)$ after the initial few epochs indicates that due to the unavailability of a proper relation between input and output, the DNN starts making its intermediate representations finer to memorize the sample-wise outcomes. Thus the expressive power of DNN is leveraged for learning the sample-specific information of the random labels thus giving high accuracy. But the limited expressive power of BNN does not allow it to do so and results in low accuracy. 

As there is not a separate phase where $I(T;X)$ decreases, there is no compression or generalization phase as such while training DNN with random labels. But in Fig. \ref{fig:MNIST_randomlabel_metrics_grad_evo}, a phase with noisy gradient is visible for DNN. Hence a noisy gradient can not be the only criterion to have a compression or generalization phase. 

While testing a DNN, in \ref{fig:MNIST_randomlabel_dynamics_test}, $I(T;Y)$ decreases in the last layer of DNN and the difference between training loss and validation loss increases over epochs (refer Fig. \ref{fig:MNIST_randomlabel_metrics_loss}), indicating clear overfitting of random labels. Hence instead of learning a relation between input and output, DNN remembers the output for each sample. This is not appreciated as for any unseen input, the network fails badly yielding high validation loss.

In the case of BNN, the training accuracy does not increase as seen in Fig. \ref{fig:MNIST_randomlabel_metrics_accuracy} and shows that a BNN does not overfit the model for the train dataset. Starting from very small values, $I(T;Y)$ and $I(T;X)$ both decrease in the last layer of BNN as shown in Fig. \ref{fig:MNIST_randomlabel_dynamics_train_DNN}. In all other intermediate layers of BNN, both $I(T;X)$ and $I(T;Y)$ increases in few initial epochs but again return back to lesser values. So, in contrast to DNN, BNN does not learn noisy labels at all and have less accuracy level in both training and validation. This indicates BNNs reliability on learning a meaningful relationship rather than just memorizing.

\end{document}